\UseRawInputEncoding
\documentclass[letterpaper, 10 pt, conference]{ieeeconf}  %

\IEEEoverridecommandlockouts                              %

\usepackage{amsmath}   %
\usepackage{amssymb}   %
\usepackage{graphicx}  %
\usepackage{booktabs}  %
\usepackage{algorithm} %
\usepackage{algorithmic}%
\usepackage{xcolor}
\usepackage{colortbl}
\usepackage{marvosym}
\usepackage[caption=false,font=footnotesize]{subfig} %
\newlength{\sensheight} %

\renewcommand{\arraystretch}{0.92}

\makeatletter
\let\NAT@parse\undefined
\g@addto@macro\normalsize{%
  \setlength{\abovedisplayskip}{4pt plus 1pt minus 2pt}%
  \setlength{\belowdisplayskip}{4pt plus 1pt minus 2pt}%
  \setlength{\abovedisplayshortskip}{3pt plus 1pt}%
  \setlength{\belowdisplayshortskip}{3pt plus 1pt minus 2pt}%
}
\makeatother

\usepackage[colorlinks]{hyperref}

\title{\LARGE \bf
AR-WAM: A Visual-Conditioned Agent-Ready World Action Model for Robotic Manipulation
}

\author{\authorblockN{Yicheng Jiang$^{1*}$, Zesen Gan$^{1*}$, Xiaobo Wang$^{2,3\text{\Letter}}$, Tianlun He$^{1}$, Chenxu Zhao$^{4}$,\\ Minghui Wu$^{4}$, Xinyue Wang$^{1}$, Jiaxu Wang$^{5}$, Junhao He$^{6}$, Jianan Wang$^{7}$, and Qiming Shao$^{1\text{\Letter}}$}%
\thanks{$^{*}$Contributed equally co-first authors, order determined by coin toss.}%
\thanks{$^{\text{\Letter}}$Corresponding authors:\newline\hspace*{2em}{\tt wangxiaobo@suat-sz.edu.cn}, {\tt eeqshao@ust.hk}.}%
\thanks{$^{1}$The Hong Kong University of Science and Technology}%
\thanks{$^{2}$Shenzhen University of Advanced Technology}%
\thanks{$^{3}$Sangfor Technologies Inc.}%
\thanks{$^{4}$Mininglamp Technology}%
\thanks{$^{5}$MMLab, The Chinese University of Hong Kong}%
\thanks{$^{6}$The University of British Columbia}%
\thanks{$^{7}$Astribot}%
}

\begin{document}

\maketitle
\thispagestyle{empty}
\pagestyle{empty}

\begin{abstract}
As AI agents become increasingly capable, agent-driven robotic control is emerging as a compelling paradigm. However, prevailing vision--language--action (VLA) models and world action models (WAMs) still rely on natural-language instructions to specify manipulation tasks, an ill-suited interface for agent-driven control: referentially ambiguous, spatially imprecise, redundant with the agent's inherent language understanding, and entangling intent with execution. We present \textbf{AR-WAM}, a visual-conditioned, agent-ready world action model that replaces language with two complementary conditions: a \emph{visual grounding prompt} (a bounding box of the target) denoting the interaction object and location, and a \emph{learnable operation token} dictating the atomic skill to execute. Our compact 0.5B-parameter model, with a frozen pretrained visual encoder and no language encoder, predicts scene evolution within compact latent states while decoding actions, exposing the policy's intent through explicit, supervisable reasoning signals. A model-agnostic compatibility layer provides three primitives (detect, execute, and query) so that local VLMs or online agent APIs can drive the policy directly, with long-horizon memory and closed-loop error recovery delegated to the agent side. On RoboTwin~2.0~\cite{chen2025robotwin2}, RMBench~\cite{chen2026rmbench}, and a real Astribot~S1~\cite{astribot2024s1} dual-arm platform, AR-WAM attains the highest average success on standard manipulation (85.7\% over the clean and randomized settings) and outperforms all baselines on memory-dependent and real-robot long-horizon tasks, improving success rates by 5.9\% and 36.7\%, respectively, while maintaining the lowest inference latency (14.1~ms). Project page is at \url{https://ar-wam.github.io/}.
\end{abstract}

\section{Introduction}
Robot control with an agent as the high-level brain and a policy model as the cerebellum is a compelling paradigm: the agent handles language understanding, task decomposition, and memory, while the policy executes precise actions. The policy side evolves along two lines: vision--language--action (VLA) models~\cite{brohan2023rt2,kim2024openvla,black2025pi0,physicalintelligence2025pi05}, which transfer web-scale semantic knowledge into continuous control, and world action models (WAMs)~\cite{du2023unipi,wu2023gr1,cheang2024gr2,hu2025vpp}, which predict scene evolution while decoding actions. Either way, the task is almost always specified by a natural-language instruction.

We argue that in agent-driven control, text conditioning is an ill-suited interface. (i)~Referential ambiguity. With several similar objects on a table, text often cannot uniquely identify the target. (ii)~Redundancy. Language understanding is what agents do best, and relearning it inside the policy wastes training compute. (iii)~Imprecision. Commands pass through the policy's own language encoder, whose imperfect grounding prevents precise steering. (iv)~Black-box entanglement. Cramming understanding, planning, memory, and action generation into one uninspectable model leaves failures unattributable between wrong intent and improper execution.

This motivates rethinking what the policy actually needs from its conditioning signal in the agent era. Open-vocabulary perception is exactly what modern vision--language models (VLMs) excel at: they localize almost arbitrary objects and return bounding boxes directly~\cite{liu2024groundingdino}, and spatial specifications have repeatedly proven more actionable than text~\cite{nasiriany2024pivot,liu2024moka,yuan2024robopoint,huang2024rekep}. We replace language with two complementary task conditions: a visual grounding prompt (a target bounding box) denotes the interaction object and location, and a learnable operation token dictates the atomic skill to execute. The agent thus retains language understanding, planning, memory, and error recovery, while the policy focuses solely on action generation.

Building on this idea, we present AR-WAM, a visual-conditioned, agent-ready world action model for robotic manipulation, with four design choices. (1)~Atomic operation decomposition. Manipulation is factorized into a compact vocabulary of atomic skills (pick, place, click, handover, etc.), each encoded by a learnable operation embedding. (2)~A lightweight latent-WAM backbone. Following the efficient latent-WAM paradigm~\cite{hu2025vpp,yang2026lilawam,chen2026lawam}, the backbone is a compact 0.5B model with a frozen pretrained visual encoder and no language encoder. A reasoning expert exposes task intent as explicit, supervisable signals, and dual-level foresight alignment keeps control future-aware. (3)~A model-agnostic compatibility layer. Three primitives (detect, execute, and query) let locally deployed VLM agents and online agent APIs drive the policy directly. (4)~Agent-side long-horizon capability and error recovery. Long-horizon tasks reuse the agent's memory, and upon a grasp failure or a dropped object the agent re-issues an updated prompt to retry, with failures staying diagnosable as intent versus execution errors.

Experiments on RoboTwin~2.0~\cite{chen2025robotwin2}, RMBench~\cite{chen2026rmbench}, and a real Astribot~S1~\cite{astribot2024s1} dual-arm platform establish that the visual prompt outperforms language conditioning under identical training, that the reasoning expert and dual-level foresight each contribute measurably to action quality, and that off-the-shelf agents drive the same policy through memory-dependent long-horizon tasks.

\section{Related Work}
\label{sec:related_work}

\subsection{VLA Models and World Action Models}

VLA models extend large-scale vision-language pretraining to end-to-end action generation. RT-2~\cite{brohan2023rt2} and OpenVLA~\cite{kim2024openvla} map web-scale vision-language knowledge to discretized action tokens, while the $\pi$ family~\cite{black2025pi0,physicalintelligence2025pi05} and X-VLA~\cite{zheng2025xvla} attach a flow-matching action expert to a pretrained VLM backbone. Despite high
performance, these all rely on language conditioning and, in the strongest cases, billion-scale backbones with massive pretraining.

World action models (WAMs) take a different route, anticipating how the scene will evolve and deriving actions from that prediction. Early WAMs reason in pixel space: UniPi plans via text-guided video generation~\cite{du2023unipi}, and GR-1/GR-2 unify video and action prediction in one sequence model~\cite{wu2023gr1,cheang2024gr2}, spending capacity on visual detail irrelevant to control. \emph{Latent WAMs} instead predict compact states, whether distilled from video models~\cite{hu2025vpp}, posed as subgoals in a foundation-model feature space~\cite{chen2026lawam}, tied to a future-informed posterior over learnable queries~\cite{luo2026beingh07}, or unified with latent action learning~\cite{bi2025motus}. Fast-WAM further shows that future prediction matters as a training signal rather than as test-time imagination~\cite{yuan2026fastwam}, and LiLa-WAM shows that a lightweight latent WAM can rival much larger models~\cite{yang2026lilawam}.

AR-WAM follows the efficient latent-WAM paradigm: with a compact 0.5B model on a frozen DINOv3 encoder~\cite{simeoni2025dinov3}, it achieves strong manipulation performance.

\subsection{Visual Grounding for Manipulation}

Existing uses of visual grounding in manipulation fall into two categories, according to the direction in which the grounding signal flows. In the first, \emph{grounding serves as part of the input}: some methods let a VLM produce spatial guidance for a low-level controller, such as ranked action candidates, affordance keypoints, or spatial constraints over keypoints or object parts~\cite{nasiriany2024pivot,liu2024moka,yuan2024robopoint,huang2024rekep,huang2024copa}. Others feed grounding directly into the policy as points, boxes, or masks~\cite{wang2026vpvla,yu2025pointvla,huang2025roboground,yu2026maskwam,zhong2025dexgraspvla}, but the grounding always comes from a fixed, built-in pipeline rather than an explicit, model-agnostic agent interface. In the second, \emph{grounding serves as an output and supervision signal}, distilled into the model through visual traces, spatially-aware encodings, or dedicated spatial-reasoning training~\cite{zheng2024tracevla,qu2025spatialvla,zhou2025roborefer}. Both lines demonstrate that spatially grounded signals are more precise and more actionable than raw language.

AR-WAM takes a more decisive step: it removes text conditioning entirely and keeps only two lightweight signals, a visual grounding box and a learnable operation token. The pair is exposed to external agents through detect--execute--query primitives, while mask prediction serves only as an auxiliary reasoning output rather than a policy input.

\subsection{Agentic Robot Control}

A line of work uses LLM- or VLM-based agents to plan over robot capabilities. Such agents score grounded proposals~\cite{ahn2022saycan}, generate executable programs~\cite{liang2023codeaspolicies}, reason over scene graphs~\cite{rana2023sayplan}, turn visual reasoning into spatial objectives for classical planners~\cite{huang2023voxposer}, or replan from language feedback~\cite{huang2022inner}. With learned policies, a memory-guided agent can steer a frozen low-level model into reliable primitives~\cite{zhang2026harnessvla}, hierarchical systems let a high-level VLM relay textual sub-task instructions to a low-level VLA~\cite{shi2025hirobot}, and dual-system designs pair a VLM planner that carries decomposition, memory, and verification with a VLA executor~\cite{liu2026goal2skill,peng2026cortex}. High-level reasoners can also detect and explain failures and replan during execution~\cite{liu2023reflect,duan2024aha}.

Two gaps remain. First, the agent-to-policy interface is text or code, which can say \emph{what} to do but not precisely \emph{where}. Second, the policy alone provides neither high-level planning nor long-horizon memory, and monolithic designs must bake them in through dedicated mechanisms~\cite{shi2025memoryvla}. AR-WAM combines both sides: the agent keeps planning, memory, and recovery and hands the policy exact image-space prompts, while the lightweight WAM focuses on reliable execution.

\section{Methodology}
\label{sec:method}

\begin{figure*}[t]
  \centering
  \includegraphics[width=1.0\linewidth]{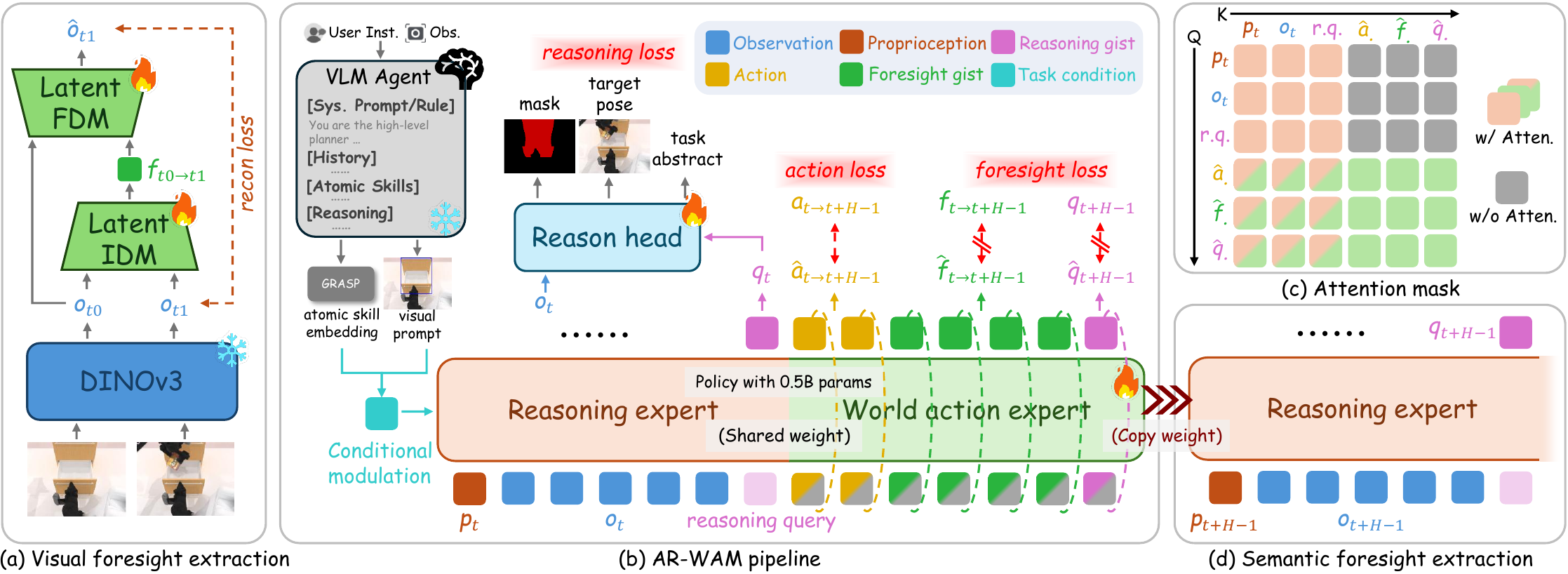}
  \caption{Overview of AR-WAM. An agent grounds the target and issues a visual prompt $g$ together with an atomic operation $s$. A compact 0.5B world action model (a 0.2B transformer shared by a reasoning expert and a world action expert, plus a frozen 0.3B vision encoder) predicts an action chunk together with dual-level foresight signals. The reasoning expert runs once per control cycle, and the world action expert then performs multi-step denoising while reusing its KV cache.}
  \vspace{-15pt}
  \label{fig:overview}
\end{figure*} 

We present AR-WAM, a \emph{visual-conditioned, agent-ready world action model} for robotic manipulation. The key design principle is to replace free-form language instructions with two complementary conditions: \emph{visual grounding prompts}, bounding boxes that spatially identify the task-relevant object, and \emph{learnable atomic-operation tokens}. This decouples \emph{where to act} (grounding, delegated to an external agent with strong open-vocabulary perception) from \emph{how to act} (control, handled by a compact policy). As a world action model, AR-WAM jointly models future states and actions, realizing the future pathway in compact latent form rather than in pixel space. Fig.~\ref{fig:overview} gives an overview.

\subsection{Task Conditioning: Atomic-Operation Embedding and Visual Prompt}
\label{sec:method:atomic}

AR-WAM organizes manipulation around a finite vocabulary of \emph{atomic operations}
$\mathcal{S} = \{$\texttt{pick}, \texttt{place}, \texttt{handover}, \texttt{click}, \texttt{pull}, \dots$\}$. Every $s \in \mathcal{S}$ is a short-horizon skill represented by a \emph{learnable atomic-operation embedding} $e_s \in \mathbb{R}^{d}$ optimized jointly with the policy. Unlike the fixed-length action chunk, a skill is delimited by execution-level events rather than a fixed horizon  (e.g., gripper opening completes a \texttt{place}), and these delimiting events coincide with those that trigger agent invocation (Sec.~\ref{sec:method:agent}). Each invocation pairs $s$ with a \emph{visual prompt}, a bounding box $g = (x_1, y_1, x_2, y_2) \in [0, 1]^4$ in normalized image coordinates that spatially specifies the target object and removes referential ambiguity. For two-arm tasks, the visual prompt extends to two boxes. The agent may further specify two auxiliary attributes: a \emph{participants} attribute $u$ (which arm acts: left, right, or both) and a \emph{style} attribute $w$ (the interaction manner, e.g., placing an object on top of another or beside it), each represented by a learnable embedding ($e_u$ and $e_w$, respectively).

The agent's decision is injected into the policy through \emph{conditional modulation} (AdaLN~\cite{peebles2023dit}): $e_s$, $e_u$, $e_w$, and the box embedding $\phi(g)$ from a lightweight MLP are fused into a \emph{task condition token} that conditions the downstream experts. All interaction behaviors executed by AR-WAM are learned skills. Auxiliary motions that require no learning (e.g., homing) are handled by classical trajectory planning at the system level.

\paragraph{Formalization}
At time step $t$, the policy receives the visual observation $o_t$, a single primary-view image, and the proprioceptive state $p_t$, and produces an action chunk $\hat{a}_{t \to t+H-1} = \{\hat{a}_{t}, \dots, \hat{a}_{t+H-1}\}$ over horizon $H$. Conditioned on the task token formed from $\phi(g)$, $e_s$, $e_u$, and $e_w$, the policy is
\begin{equation}
    \hat{a}_{t \to t+H-1} \sim \pi_\theta\!\left(\,\cdot \mid o_t,\ p_t,\ \phi(g),\ e_s,\ e_u,\ e_w\,\right),
    \label{eq:policy}
\end{equation}
where the parameters $\theta$ are shared across all atomic operations, so that data from different skills reinforce a shared visuomotor representation and the model remains a single deployable artifact. 
Here $\pi_\theta$ abstracts the entire agent-facing policy, instantiated by the reasoning expert $f_\theta^{\mathrm{reas}}$ (Sec.~\ref{sec:method:reasoning}) and the world action expert $f_\theta^{\mathrm{wam}}$ (Sec.~\ref{sec:method:wam})

\subsection{Reasoning Expert and Foresight Gist}
\label{sec:method:reasoning}

The reasoning expert grounds the model's \emph{task understanding} before actions are generated: a set of \emph{learnable reasoning query tokens} $q^{\mathrm{reas}}$ probes the observation and the task condition, producing a \emph{reasoning token} $q_t$:
\begin{equation}
    q_t = f_{\theta}^{\mathrm{reas}}\!\left(\,E(o_t),\ p_t,\ \phi(g),\ e_s,\ e_u,\ e_w,\ q^{\mathrm{reas}}\,\right),
    \label{eq:reasoning}
\end{equation}
where $E(\cdot)$ is the frozen DINOv3 visual encoder~\cite{simeoni2025dinov3}. A lightweight \emph{reason head} decodes $q_t$ into three explicit, supervisable predictions of the current step (cf. Fig.~\ref{fig:overview}):
\begin{equation}
    (\hat{m}_t,\ \hat{T}_t,\ \hat{c}_t) = D^{\mathrm{reas}}(q_t),
    \label{eq:reasonhead}
\end{equation}
namely (i)~the \emph{mask} $\hat{m}_t$ of the boxed object, which sharpens the coarse box into a pixel-precise target; (ii)~the end-effector poses $\hat{T}_t$ of both arms when the executing skill terminates; and (iii)~a \emph{task abstract} $\hat{c}_t$, a structured one-hot description of the current step (e.g., which arm to use, and how to grasp and place). Unlike the text-based reasoning common in VLAs, these predictions expose the model's action intent in supervisable form: they are trained against the ground-truth annotations $(m_t, T_t, c_t)$ by the reasoning loss
\begin{equation}
    \mathcal{L}_{\mathrm{reas}} = \mathrm{BCE}(\hat{m}_t, m_t) + \mathrm{CE}(\hat{c}_t, c_t) + \lVert \hat{T}_t - T_t \rVert^2,
    \label{eq:reasloss}
\end{equation}
and serve as failure diagnostics (Sec.~\ref{sec:method:agent}).

\paragraph{Foresight-gist extraction}

To supply a compact, future-aware training target, we pre-train an \emph{offline} foresight-gist extractor on demonstration sequences: the frozen encoder $E$ maps two endpoint frames $o_{t_0}$ and $o_{t_1}$ ($t_0 < t_1$) of a clip into latent frames, and a latent inverse-dynamics module compresses their temporal evolution into a \emph{foresight gist}, with no actions or intermediate frames:
\begin{align}
  f_{t_0 \to t_1} = G^{\mathrm{IDM}}(E(o_{t_0}), E(o_{t_1})).
  \label{eq:gist}
\end{align}
A latent forward-dynamics module then reconstructs the future latent frame from the current frame and the gist,
\begin{align}
  \hat{e}_{t_1} = G^{\mathrm{FDM}}(E(o_{t_0}), f_{t_0 \to t_1}),
  \label{eq:fdm}
\end{align}
and the pair is trained with a \emph{latent} reconstruction loss:
\begin{align}
  \mathcal{L}_{\mathrm{recon}} = \lVert \hat{e}_{t_1} - E(o_{t_1}) \rVert^2.
  \label{eq:recon}
\end{align}
The gist captures \emph{how the scene will evolve} in compact latent form. Computed from future observations, it serves only as a training-time target and is never used at inference.

\subsection{World Action Expert: Dual-Level Future Alignment}
\label{sec:method:wam}

The world action expert generates executable actions while being explicitly aligned with the future at two levels, so that its decisions are shaped by their anticipated consequences. It shares its transformer with the reasoning expert, invoked with different input tokens, letting task understanding directly shape control. Given $E(o_t)$, $p_t$, and the task condition token as a shared condition, it jointly denoises three output modalities by flow matching with a single denoiser $f_\theta^{\mathrm{wam}}$: from noise $z^0 \sim \mathcal{N}(\mathbf{0}, \mathbf{I})$ sampled independently per modality and one shared flow time $\tau$, each prediction is $\hat{z} = f_\theta^{\mathrm{wam}}(z^\tau, \tau)$ at the interpolated state $z^\tau = (1-\tau)\,z^0 + \tau z^1$, $z \in \{a, f, q\}$:
\begin{itemize}
    \item an \textbf{action chunk} $\hat{a}_{t \to t+H-1}$ (Eq.~\eqref{eq:policy}), with $a^1 = a_{t \to t+H-1}$ the demonstrated chunk:
    \begin{align}
        \mathcal{L}_{\mathrm{act}} = \mathbb{E}\,\big\lVert f_\theta^{\mathrm{wam}}(a^\tau, \tau) - (a^1 - a^0) \big\rVert^2, &&
        \label{eq:actloss}
    \end{align}
    \item a \textbf{visual foresight} prediction $\hat{f}_{t \to t+H-1}$, an anticipated representation of \emph{how the scene will look}, aligned to the foresight gist extracted offline from the two endpoint frames, $f_{t \to t+H-1} = G^{\mathrm{IDM}}(E(o_t), E(o_{t+H-1}))$ (Sec.~\ref{sec:method:reasoning}):
    \begin{align}
        \mathcal{L}_{\mathrm{vis}} = \big\lVert f_\theta^{\mathrm{wam}}(f^\tau, \tau) - \mathrm{sg}(f_{t \to t+H-1}) \big\rVert^2, &&
        \label{eq:foresight}
    \end{align}
    \item a \textbf{semantic foresight} prediction $\hat{q}_{t+H-1}$, an anticipated representation of \emph{what the scene will mean}, aligned to the reasoning token $q_{t+H-1}$ that the weight-sharing reasoning backbone produces from the \emph{future} observation $(o_{t+H-1}, p_{t+H-1})$ (Eq.~\eqref{eq:reasoning}):
    \begin{align}
        \mathcal{L}_{\mathrm{sem}} = \big\lVert f_\theta^{\mathrm{wam}}(q^\tau, \tau) - \mathrm{sg}(q_{t+H-1}) \big\rVert^2, &&
        \label{eq:semforesight}
    \end{align}
\end{itemize}
where $\mathrm{sg}(\cdot)$ denotes the stop-gradient operator: the foresight branches shape the shared representation without back-propagating into the observation-encoding path. The action branch regresses the interpolation velocity, while the two foresight branches regress clean-endpoint estimates of their targets. The semantic target $q_{t+H-1}$ is computed online by the same backbone under a stop gradient rather than by a frozen copy, and since the reasoning token is explicitly supervised by the reason head, this bootstrapped target does not suffer from representation collapse. A block attention mask over the token groups $\{E(o_t), p_t, q^{\mathrm{reas}}, \hat{a}, \hat{f}, \hat{q}\}$ lets the action and foresight branches condition on the reasoning output without contaminating it: the reasoning pass and the first denoising step share one forward per control cycle, and the remaining steps reuse the cached keys and values. At inference, all three start from independent Gaussian noise and are jointly refined over a fixed number of ODE integration steps.

\subsection{Training Objective}
\label{sec:method:objective}

The complete model is trained end-to-end with the joint objective
\begin{equation}
    \mathcal{L}(\theta) = \mathcal{L}_{\mathrm{act}} + \lambda_{\mathrm{reas}}\,\mathcal{L}_{\mathrm{reas}} + \lambda_{\mathrm{vis}}\,\mathcal{L}_{\mathrm{vis}} + \lambda_{\mathrm{sem}}\,\mathcal{L}_{\mathrm{sem}},
    \label{eq:loss}
\end{equation}
where the four terms are defined in Eqs.~\eqref{eq:actloss}, \eqref{eq:reasloss}, \eqref{eq:foresight}, and \eqref{eq:semforesight}, and the $\lambda$ coefficients balance them.

\paragraph{Data}
Training data, detailed in Sec.~\ref{sec:exp-sim}, mixes simulation data from RoboTwin~2.0~\cite{chen2025robotwin2} and RMBench~\cite{chen2026rmbench}, whose ground-truth object states are converted into bounding-box visual prompts $g$.

\paragraph{Visual-prompt augmentation}
Since agent-issued boxes may be loose or slightly misplaced at deployment, we randomly jitter the box center, height, and width during training, so the policy tolerates imprecise boxes instead of overfitting to exact coordinates.
 
\subsection{Agent Compatibility Layer and Failure Recovery}
\label{sec:method:agent}

We model the agent as a \emph{model plus harness}~\cite{trivedy2026harness,wang2023agentsurvey}: a large (vision-)language model wrapped in a system prompt and rules, callable tools, interaction history, and a control loop. Perception in such agents is \emph{native}, with vision--language models emitting structured grounding outputs such as bounding boxes directly. Given its history $\mathcal{H}_t$ and the current observation, the agent issues the next command $(y_t, s_t, u_t, w_t) = \mathcal{A}(\mathcal{H}_t, o_t)$ through the three primitives introduced below.

\paragraph{Event-triggered agent invocation}
The agent is not called at every control step. Instead, each invocation is triggered by one of four events monitored during the policy rollout: (i) cumulative end-effector displacement since the last invocation exceeds a threshold, (ii) gripper closure, (iii) gripper opening, and (iv) a fixed time interval elapses since the last invocation. At each trigger, the agent receives structured feedback: the trigger reason, the current gripper state, and the observation at the trigger moment.

\paragraph{Agent context and output}
The agent's context consists of a global system prompt and rules, the task instruction, the first frame, the atomic-skill list, and the history of policy feedback, agent reasoning and command pairs ($\mathcal{H}$ in Algorithm~\ref{alg:interface}). Conditioned on it, the agent outputs a reasoning segment followed by the next command.

AR-WAM exposes a \emph{model-agnostic interface} of three primitives: \texttt{detect}, \texttt{execute}, and \texttt{query}, so that any agent can operate the robot without knowing the policy internals. Algorithm~\ref{alg:interface} shows the resulting closed loop.

\begin{algorithm}[t]
\caption{Closed-loop agent--policy interaction}
\label{alg:interface}
\begin{algorithmic}[1]
\STATE \textbf{Input:} task instruction, first observation $o_0$
\STATE $\mathcal{H} \leftarrow \{$system prompt, task instruction, $o_0$, atomic-skill list $\mathcal{S}$$\}$
\REPEAT
  \STATE $(y, s, u, w) \leftarrow \textsc{AgentReason}(\mathcal{H})$ \COMMENT{reason over the history and issue the next command: grounding query $y$ and skill $s \in \mathcal{S}$}
  \STATE $g \leftarrow \texttt{detect}(y)$ \COMMENT{agent-side grounding}
  \STATE $\xi \leftarrow \texttt{execute}(g, s, u, w)$ \COMMENT{roll out $\pi_\theta$ until a trigger event fires}
  \STATE $b \leftarrow \texttt{query}(\xi)$ \COMMENT{trigger reason, gripper state, observation}
  \STATE $\mathcal{H} \leftarrow \mathcal{H} \cup \{(y, s, g, b)\}$ \COMMENT{command and feedback enter the agent's memory}
\UNTIL{the agent judges the task complete from $\mathcal{H}$}
\end{algorithmic}
\end{algorithm}

The three primitives partition responsibility cleanly: $\texttt{detect}(y) \rightarrow g$ runs entirely on the agent side: the query $y$ (e.g., ``the red mug'') is grounded in the current observation using the agent's own grounding capability, and only the box ever reaches the policy, which itself contains no language pathway. $\texttt{execute}(g, s, u, w) \rightarrow \xi$ rolls out the policy $\pi_\theta$ (Eq.~\eqref{eq:policy}) until one of the trigger conditions above fires and returns the execution trace $\xi$. $\texttt{query}(\xi) \rightarrow b$ packages this outcome feedback for the agent's next reasoning step.

This design makes the agent fully interchangeable: switching from a locally deployed VLM to an online API model changes only the agent-side grounding and reasoning abilities, never the policy.

\paragraph{Long-horizon tasks and memory}
\label{sec:method:memory}
AR-WAM deliberately keeps the policy $\pi_\theta$ \emph{Markovian}: each decision depends only on the current observation and conditioning $(g, s, u, w)$, with no recurrent state inside the policy. All long-horizon structure, such as tracking completed sub-goals, failures, and next steps, is carried by the agent's own history and memory. AR-WAM thereby avoids the dedicated memory architectures that monolithic reactive policies require~\cite{shi2025memoryvla}, and extends the achievable horizon by extending the agent's context rather than retraining the policy.

\begin{figure}[ht]
  \centering
  \includegraphics[width=\columnwidth]{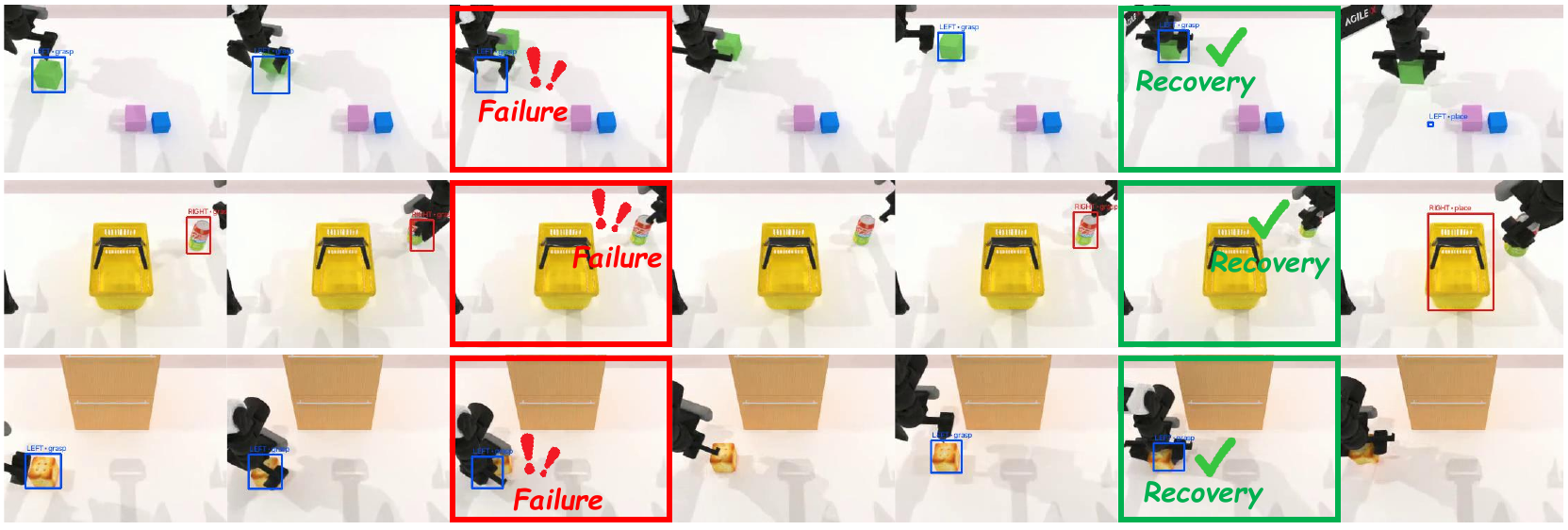}
  \caption{Failure recovery through the agent--policy loop.}
  \label{fig:failure-recovery}
\end{figure}

\paragraph{Failure detection and recovery}
\label{sec:method:recovery}
Failures such as unsuccessful grasps or dropped objects are handled by the closed loop above (Fig.~\ref{fig:failure-recovery}) rather than by hardwired recovery behaviors: at each invocation, the agent judges subtask completion, and on failure re-runs \texttt{detect} for an updated prompt $g'$ (e.g., re-localizing a dropped object) and re-invokes $\texttt{execute}(g', s, u, w)$ under a fixed retry budget. Because retries enter through the same interface, recovery composes naturally with agent-side memory, and the policy's decoded intent (Sec.~\ref{sec:method:reasoning}) helps diagnose whether a failure stems from a wrong intent or from improper execution.

\section{Experiments}
\label{sec:experiments}

Our experiments are organized around three questions:

\textbf{(RQ1)} Can visual grounding prompts replace, or even outperform, textual instructions as task conditioning?

\textbf{(RQ2)} Does the model-agnostic agent compatibility layer enable diverse VLM/LLM agents to drive the same policy on long-horizon and memory-dependent tasks?

\textbf{(RQ3)} How do the conditioning and modulation choices affect AR-WAM's manipulation efficiency and performance?

\subsection{Simulation Benchmarks}
\label{sec:exp-sim}

\textbf{Setup.}
We evaluate AR-WAM on dual-arm manipulation tasks from RoboTwin~2.0~\cite{chen2025robotwin2} and on RMBench~\cite{chen2026rmbench}, with the policy trained jointly on data from both benchmarks. From RoboTwin~2.0, we select ten long-horizon tasks that require bimanual coordination and compose rich combinations of the atomic skills supported by AR-WAM (\texttt{pick}, \texttt{place}, \texttt{handover}, \dots) to primarily probe manipulation precision. To factor out the influence of any particular VLM, instructions are issued by a state-machine-based benchmark expert equipped with an object detector.

RMBench, in contrast, comprises memory-dependent tasks that test reasoning and memory, i.e., the cooperation between the policy and the agent. There, instructions are provided by a locally deployed Qwen3.8-27B~\cite{qwen2026qwen38} and the online API Kimi-K3~\cite{kimi2026kimik3}, both of which are open-source and natively multimodal.

We re-generate the demonstration data with RoboTwin~2.0 (50 clean and 500 randomized demonstrations per task) and RMBench (50 demonstrations per task) to train the model in the main experiments.
During data generation, a homing action that returns the arm to a rest pose is inserted between consecutive operations whose execution may occlude the manipulated object, so that the outcome of every subtask remains clearly observable and the agent can reliably judge its success or failure.
All methods are evaluated on an identical, fixed set of initial-state seeds.

\textbf{Baselines.}
We select baselines from the model families most related to AR-WAM: language-conditioned VLAs ($\pi_{0.5}$~\cite{physicalintelligence2025pi05}, X-VLA~\cite{zheng2025xvla}), world action models (Fast-WAM~\cite{yuan2026fastwam}), and latent world action models (Motus~\cite{bi2025motus}, LiLa-WAM~\cite{yang2026lilawam}). On RoboTwin~2.0, we compare against all five. On RMBench, we compare against $\pi_{0.5}$, Fast-WAM, and LiLa-WAM, together with three memory-augmented policies: Mem-0, the reference policy released with the benchmark~\cite{chen2026rmbench}, and Goal2Skill~\cite{liu2026goal2skill} and Cortex~\cite{peng2026cortex}, two recent hierarchical systems that, like AR-WAM, pair a memory-aware high-level VLM agent with a low-level visuomotor policy.
For baselines with officially released weights, we directly adopt the public checkpoints. All other baselines are retrained on the same demonstrations as AR-WAM.

\textbf{Results.}
Table~\ref{tab:main} reports per-task and average success rates. AR-WAM attains the best average success rate in the random setting ($84.2\%$) and stays within 0.6\% of the strongest baseline in the clean setting ($87.2\%$ vs.\ $87.8\%$), achieving higher success than the evaluated language-conditioned baselines under this evaluation protocol (RQ1).

\begin{table}[ht]
  \centering
  \caption{Success rates (\%) on ten RoboTwin~2.0 tasks over 50 trials per task, per setting.
  C: clean setting; R: randomized setting. \textbf{Bold}: best; \underline{underline}: second best.}
  \label{tab:main}
  \setlength{\tabcolsep}{2.2pt}
  \renewcommand{\arraystretch}{1.05}
  \resizebox{\columnwidth}{!}{%
  \begin{tabular}{lcccccccccc>{\columncolor{blue!5}}c>{\columncolor{blue!5}}c}
    \toprule
    & \multicolumn{2}{c}{Pi0.5} & \multicolumn{2}{c}{X-VLA} & \multicolumn{2}{c}{Motus}
    & \multicolumn{2}{c}{Fast-WAM} & \multicolumn{2}{c}{LiLa-WAM} & \multicolumn{2}{>{\columncolor{blue!5}}c}{AR-WAM} \\
    \cmidrule(lr){2-3}\cmidrule(lr){4-5}\cmidrule(lr){6-7}\cmidrule(lr){8-9}\cmidrule(lr){10-11}\cmidrule(lr){12-13}
    Task & C & R & C & R & C & R & C & R & C & R & C & R \\
    \midrule
    Blocks Ranking Size & 49 & 26 & 67 & 74 & 75 & 63 & \underline{94} & \textbf{98} & 92 & 88 & \textbf{100} & \underline{94} \\
    Handover Block      & 66 & 57 & 73 & 37 & 86 & 73 & \textbf{99} & 80 & 96 & \underline{90} & \underline{98} & \textbf{92} \\
    Handover Mic        & \underline{98} & 97 & 0 & 0 & 78 & 63 & \textbf{100} & \textbf{100} & \textbf{100} & \underline{98} & \textbf{100} & \textbf{100} \\
    Hanging Mug         & 18 & 17 & 23 & 27 & 38 & 38 & \textbf{65} & \underline{56} & \underline{56} & 44 & 54 & \textbf{62} \\
    Place Can Basket    & 62 & 62 & 49 & 52 & \underline{81} & \underline{76} & 72 & 67 & 80 & 72 & \textbf{90} & \textbf{82} \\
    Place Dual Shoes    & 75 & 75 & 79 & \textbf{88} & \textbf{93} & \underline{87} & \underline{88} & \textbf{88} & 60 & 54 & 86 & 78 \\
    Put Bottles Dustbin & 84 & 79 & 74 & 77 & 81 & 79 & \underline{93} & 82 & 92 & \textbf{94} & \textbf{96} & \underline{92} \\
    Put Object Cabinet  & 80 & 79 & 46 & 48 & 88 & 71 & \textbf{94} & \underline{82} & \underline{92} & \textbf{92} & 78 & 70 \\
    Scan Object         & 72 & 65 & 14 & 36 & 67 & 66 & \textbf{96} & \underline{86} & \underline{94} & \textbf{90} & 76 & \underline{86} \\
    Stack Bowls Three   & 77 & 71 & 76 & \underline{86} & 79 & \textbf{87} & 77 & \underline{86} & \underline{88} & 82 & \textbf{94} & \underline{86} \\
    \midrule
    Average & 68.1 & 62.8 & 50.1 & 52.5 & 76.6 & 70.3 & \textbf{87.8} & \underline{82.5} & 85.0 & 80.4 & \underline{87.2} & \textbf{84.2} \\
    \bottomrule
  \end{tabular}%
  }
  \vspace{-5pt}
\end{table}

\subsection{Conditioning, Modulation, and Component Ablations}
\label{sec:exp-condition}

To answer RQ1 and RQ3, we ablate five design axes on the RoboTwin~2.0 suite under an identical backbone, data budget, and training schedule: the \emph{conditioning modality}, the \emph{reasoning} and \emph{foresight} components, and the \emph{foresight objective} (Table~\ref{tab:ablation}), together with the \emph{modulation mechanism} (Table~\ref{tab:sensitivity}(b)).

\begin{table}[!t]
  \centering
  \caption{Ablation on condition, reasoning and foresight components.
  All variants use AdaLN modulation. VP: visual prompt. Objective: foresight
  prediction target.
  Params (B). Avg SR: average success rate (\%).}
  \label{tab:ablation}
  \footnotesize
  \setlength{\tabcolsep}{4pt}
  \resizebox{\columnwidth}{!}{%
  \begin{tabular}{lccccc}
    \toprule
    Condition & Reasoning & Foresight & Objective & Params & Avg SR \\
    \midrule
    Text      & $\checkmark$ & $\checkmark$ & Gist  & 0.58 & 74.2 \textcolor{red}{($-$13.0)} \\
    VP        & $\times$     & $\checkmark$ & Gist  & 0.49 & 79.4 \textcolor{red}{($-$7.8)} \\
    VP        & $\checkmark$ & $\times$     & --    & 0.50 & 78.6 \textcolor{red}{($-$8.6)} \\
    VP + Text & $\checkmark$ & $\checkmark$ & Gist  & 0.67 & 84.6 \textcolor{red}{($-$2.6)} \\
    VP        & $\checkmark$ & $\checkmark$ & Patch & 0.51  & 76.4 \textcolor{red}{($-$10.8)} \\
    \midrule
    VP        & $\checkmark$ & $\checkmark$ & Gist  & 0.50 & 87.2 \\
    \bottomrule
  \end{tabular}%
  }
  \vspace{-5pt}
\end{table}

\paragraph{Conditioning and modulation}
Replacing the visual prompt with text instructions (Table~\ref{tab:ablation}) drops average success by $13.0\%$, directly answering RQ1: under identical training, visual grounding prompts are a strictly stronger conditioning signal than language. Adding text on top of the visual prompt does not help either ($-2.6\%$), indicating that language is redundant once spatial grounding is available.

Replacing AdaLN with cross-attention drops average success by $13.8\%$ (Table~\ref{tab:sensitivity}(b)). To understand \emph{why}, we measure each variant's reliance on every condition group (Table~\ref{tab:sensitivity}(a)): we permute one group at a time (the skill embedding alone, the full \emph{semantic bundle} of skill, participants, and style encodings jointly, the proprioceptive state, or all jointly) at matched optimization steps and record the induced change in predicted action velocity, with flow time, noise, and RNG state fixed.

Total reliance is comparable but placed differently: cross-attention depends $2.3\times$ more on the proprioceptive state ($35.6\%$ vs.\ $15.5\%$), whereas AdaLN depends $2.1\times$ more on the semantic bundle ($31.5\%$ vs.\ $15.3\%$). Such proprioception-heavy reliance is consistent with the shortcut-learning failure mode of imitation policies~\cite{zhao2025statefree,lu2026gap}: the state signal correlates strongly with expert actions in-distribution and offers a faster route to loss reduction, but carries no task semantics, so leaning on it bypasses the semantic condition and undermines generalization.

\begin{table}[!t]
  \centering
  \caption{Comparison of the two modulation mechanisms. 
  Params (B). Avg SR: average success rate (\%).}
  \vspace{-12pt}
  \label{tab:sensitivity}
  \settoheight{\sensheight}{\includegraphics[width=0.575\linewidth]{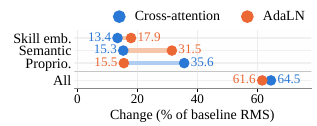}}%
  \subfloat[Condition sensitivity]{%
    \includegraphics[width=0.575\linewidth]{figures/condition_sensitivity.pdf}%
  }\hfill
  \subfloat[Success rates]{%
    \begin{minipage}[b][\sensheight][c]{0.40\linewidth}
      \centering
      \scriptsize
      \setlength{\tabcolsep}{2.2pt}
      \renewcommand{\arraystretch}{1.05}
      \resizebox{\linewidth}{!}{%
      \begin{tabular}{@{}lcc@{}}
        \toprule
        Modulation & Params & Avg SR \\
        \midrule
        Cross & 0.44 & 73.4 \textcolor{red}{($-$13.8)} \\
        AdaLN & 0.50 & 87.2 \\
        \bottomrule
      \end{tabular}%
      }
    \end{minipage}%
  }
  \vspace{-3pt}
\end{table}

\begin{figure}[!t]
  \centering
  \includegraphics[width=1.0\linewidth]{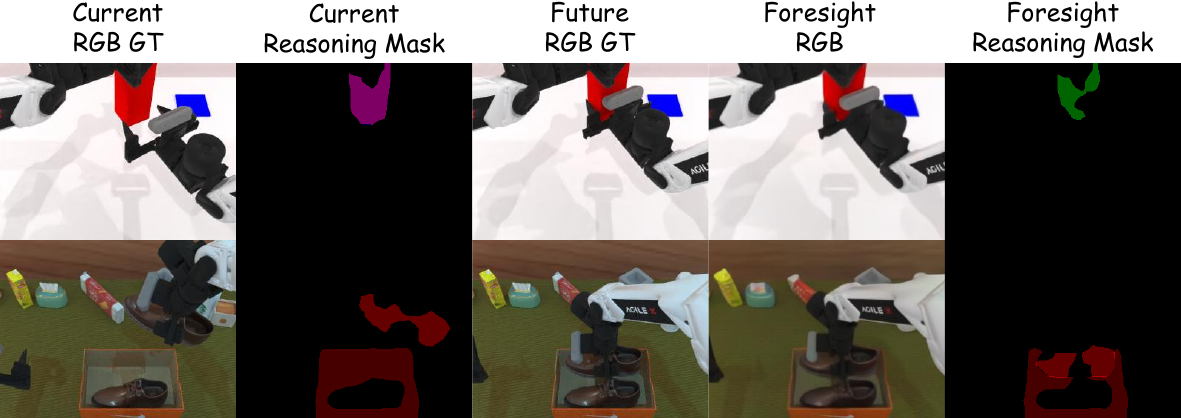}
  \caption{Foresight prediction. Given the current frame and the task condition, AR-WAM predicts the latent foresight of the future scene, decoded here for visualization, together with the target masks decoded by the reason head for the current and the foreseen scene.}
  \label{fig:predict}
  \vspace{-5pt}
\end{figure}

\begin{figure}[!t]
  \centering
  \includegraphics[width=1.0\linewidth]{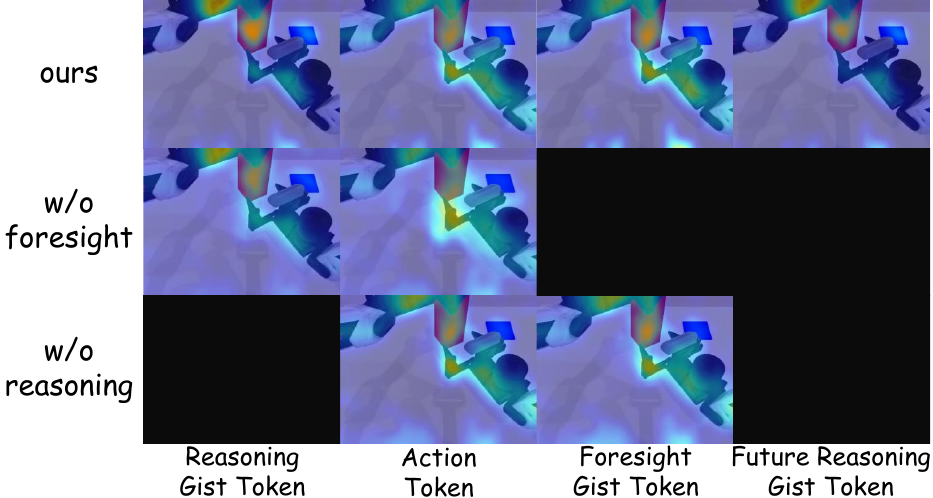}
  \caption{Attention maps of different token groups under three setups (full model, without foresight, without reasoning).}
  \label{fig:mask}
  \vspace{-5pt}
\end{figure}

\paragraph{Reasoning and foresight}
Removing the reasoning expert costs $7.8\%$ of average success and removing the foresight alignment $8.6\%$ (Table~\ref{tab:ablation}), confirming that both explicit task understanding and future-aware latent targets contribute to action quality (RQ3). Replacing the compact gist objective with raw future patch features (Patch) drops average success by 10.8\%, below even the foresight-free variant, suggesting that the gist's compression is what makes the foresight signal useful. Fig.~\ref{fig:predict} visualizes the decoded foresight predictions against the ground-truth future frames, and Fig.~\ref{fig:mask} compares the attention maps of different token groups under three setups: reasoning concentrates attention on the object to be manipulated and the target, while foresight shifts attention toward regions where motion will occur, so that together the model attends to both the manipulated object and the scene dynamics.

\subsection{Agent-Driven Long-Horizon Evaluation on RMBench}
\label{sec:exp-rmbench}

To answer RQ2, we evaluate AR-WAM on RMBench~\cite{chen2026rmbench}, pairing the same policy with Qwen3.8-27B and Kimi-K3 as the agents through the model-agnostic interface of Sec.~\ref{sec:method:agent}. Table~\ref{tab:rmbench} reports per-task and average success rates.

\begin{table}[!t]
  \centering
  \caption{Success rates (\%) on RMBench over 50 trials per task.
  AR-K / AR-Q: AR-WAM integrated with Kimi-K3~\cite{kimi2026kimik3} / Qwen3.8-27B~\cite{qwen2026qwen38} as the agent.
  --: not applicable / not reported. \textbf{Bold}: best; \underline{underline}: second best.}
  \label{tab:rmbench}
  \footnotesize
  \setlength{\tabcolsep}{2.4pt}
  \resizebox{\columnwidth}{!}{%
  \begin{tabular}{lcccccc>{\columncolor{blue!5}}c>{\columncolor{blue!5}}c}
    \toprule
    Task & Pi0.5 & Fast-WAM & Goal2Skill & Mem-0 & Cortex & LiLa-WAM & AR-K & AR-Q \\
    \midrule
    Observe and Pick Up & 9  & 0 & 8  & 4  & 14  & 10 & \underline{44} & \textbf{84} \\
    Rearrange Blocks    & 13 & 0 & 38 & \underline{89} & \textbf{100} & 20 & 62 & 86 \\
    Put Back Block      & 11 & 0 & -- & \underline{90} & \textbf{100} & 14 & 58 & 88 \\
    Swap Blocks         & 24 & 0 & -- & 67 & \textbf{99}  & 10 & 68 & \underline{82} \\
    Swap T              & 15 & 7 & -- & 14 & \textbf{63}  & \underline{18} & \underline{18} & 16 \\
    \midrule
    $M(1)$ Tasks Average & 14.4 & 1.4 & 23.0 & 52.8 & \textbf{75.2} & 14.4 & 50.0 & \underline{71.2} \\
    \midrule
    Battery Try         & 16 & 20 & \underline{46} & 28 & 37 & 24 & 38 & \textbf{52} \\
    Blocks Ranking Try  & 6  & 26 & \textbf{60} & 18 & -- & 30 & 18 & \underline{34} \\
    Cover Blocks        & 0  & 0  & -- & 68 & -- & 0 & \underline{72} & \textbf{78} \\
    Press Button        & 0  & 0  & 10 & 0  & 20 & 0 & \underline{82} & \textbf{90} \\
    \midrule
    $M(n)$ Tasks Average & 5.5 & 11.5 & 38.7 & 28.5 & 28.5 & 13.5 & \underline{52.5} & \textbf{63.5} \\
    \midrule
    Total Average       & 10.4 & 5.9 & 32.4 & 42.0 & \underline{61.9} & 14.0 & 51.1 & \textbf{67.8} \\
    \bottomrule
  \end{tabular}%
  }
\end{table}

On $M(n)$ tasks, AR-WAM with either agent substantially outperforms all memory-augmented baselines ($52.5\%$ and $63.5\%$ vs.\ $38.7\%$ for the strongest, Goal2Skill), which we attribute to clearly delegating memory management to the upstream agent rather than entangling it within the policy.

On $M(1)$ tasks, however, AR-WAM trails Cortex ($75.2\%$). These tasks involve smaller objects, for which the bounding boxes produced natively by the VLM are not always accurate, introducing occasional manipulation errors. AR-Q surpasses AR-K ($71.2\%$ vs.\ $50.0\%$ on $M(1)$) despite AR-K's far larger agent model, suggesting that once both agents reason well enough to handle the memory component, grounding accuracy becomes the bottleneck. On the ``Swap T'' task, AR-WAM does not improve over LiLa-WAM, indicating that bounding boxes, as a form of visual grounding, are not yet fine-grained enough for tasks that demand strict spatial details.

\subsection{Real-Robot Evaluation}
\label{sec:exp-real}

\begin{figure}[!t]
  \centering
  \includegraphics[width=1.0\linewidth]{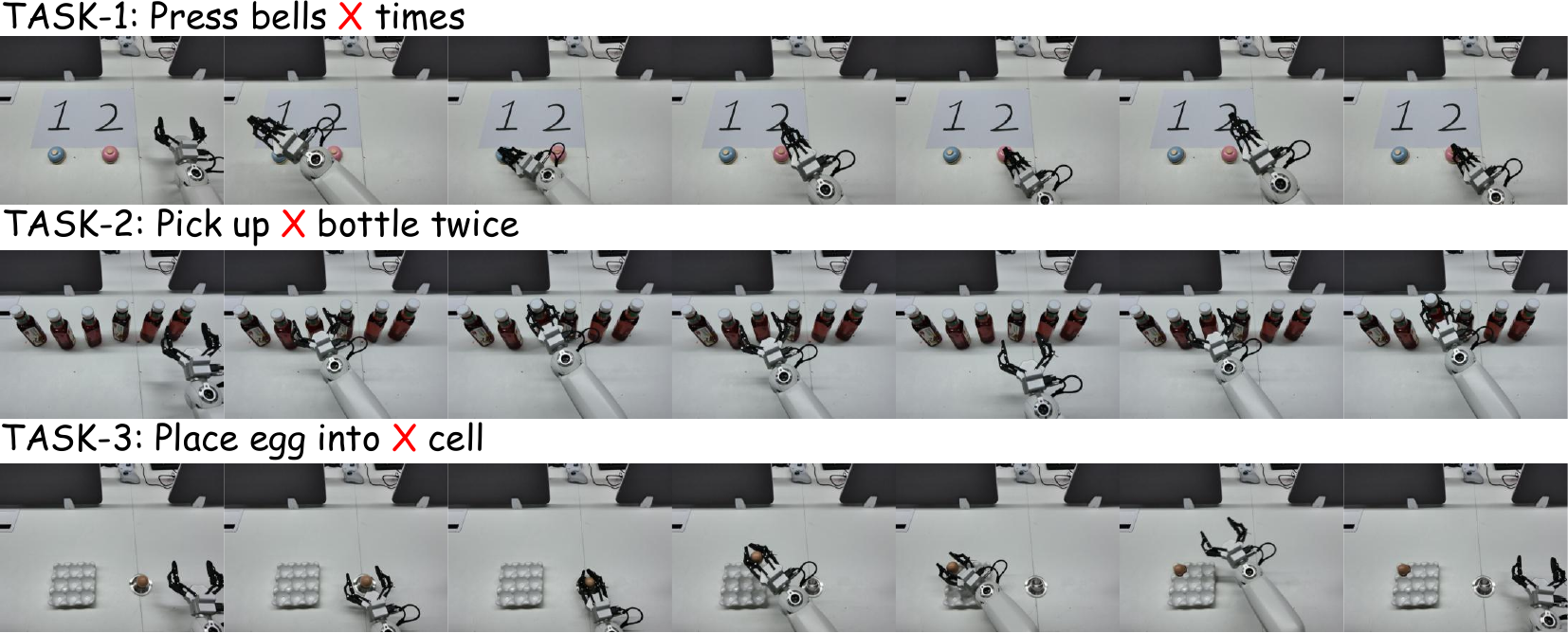}
  \caption{Real-robot rollouts of the three tasks on the Astribot~S1~\cite{astribot2024s1} dual-arm platform: pressing bells a specified number of times, picking up a specified bottle twice, and placing an egg into a specified cell.}
  \label{fig:real}
\end{figure}

We design three challenging real-robot tasks on the Astribot~S1~\cite{astribot2024s1} dual-arm platform (Fig.~\ref{fig:real}), all long-horizon and requiring agent-side reasoning and memory, and evaluate AR-WAM driven by a locally deployed Qwen3.8-27B~\cite{qwen2026qwen38} as the agent. For each task, we finetune the simulation-trained policy on $100$ expert demonstrations collected on the real platform. AR-WAM achieves the highest success on all three tasks (Table~\ref{tab:realrobot}). These results demonstrate that AR-WAM is not only effective in benchmark settings but practically viable for real-world robotic deployment.

\begin{table}[ht]
  \centering
  \caption{Success rates (\%) on three challenging real-robot tasks over 20 trials per task. \textbf{Bold}: best.}
  \label{tab:realrobot}
  \footnotesize
  \setlength{\tabcolsep}{4pt}
  \resizebox{0.85\columnwidth}{!}{%
  \begin{tabular}{lcc>{\columncolor{blue!5}}c}
    \toprule
    Task & Pi0.5 & LiLa-WAM & \textbf{AR-WAM} \\
    \midrule
    Press bells X times & 5 & 10 & \textbf{70} \\
    Pick up X bottle twice & 45 & 40 & \textbf{65} \\
    Place egg into X cell & 55 & 45 & \textbf{80} \\
    \midrule
    Average & 35.0 & 31.7 & \textbf{71.7} \\
    \bottomrule
  \end{tabular}%
  }
\end{table}

\subsection{Efficiency Analysis}
\label{sec:exp-efficiency}
To assess the efficiency aspect of RQ3, we compare AR-WAM with policies of a similar mixture-of-transformers structure (Table~\ref{tab:latency}), whose computation splits into a reasoning part and an action-denoising part, so that the end-to-end latency from receiving the current observation to producing an action decomposes into three stages: (i)~\emph{input processing}, which encodes the task condition and image inputs into tokens, (ii)~\emph{KV-cache construction}, typically performed together with the first ODE (denoising) step, and (iii)~\emph{per-step ODE time}, the cost of each remaining denoising step. All latencies are measured on a single RTX 4090 (24~GB) in bf16 precision with identical input data across all methods, and each figure is averaged over repeated runs after several warm-up iterations. AR-WAM attains the lowest end-to-end latency (14.09~ms per chunk), $4.4\times$ faster than $\pi_{0.5}$ and $22.7\times$ faster than Fast-WAM, benefiting from the shared reasoning-denoising forward and KV-cache reuse.

\begin{table}[ht]
  \centering
  \caption{Per-stage inference latency (ms).
  Encoder \& Misc.: visual encoding and other overheads;
  ODE: latency per denoising step, with the number of steps in superscript. \textbf{Bold}: fastest; \underline{underline}: second fastest.}
  \label{tab:latency}
  \footnotesize
  \setlength{\tabcolsep}{4pt}
  \resizebox{\columnwidth}{!}{%
  \begin{tabular}{lcccc}
    \toprule
    Method & Encoder \& Misc. & KV Cache & ODE & End-to-end \\
    \midrule
    Pi0.5    & 19.74 & \underline{23.36} & 2.11$^{(\times 9)}$ & 62.09 \\
    Fast-WAM & 27.55 & 52.56 & 26.58$^{(\times 9)}$ & 319.33 \\
    LiLa-WAM & \underline{4.77} & -- & \underline{1.02}$^{(\times 10)}$ & \underline{14.97} \\
    \midrule
    AR-WAM & \textbf{4.41} & \textbf{3.74} & \textbf{0.66}$^{(\times 9)}$ & \textbf{14.09} \\
    \bottomrule
  \end{tabular}%
  }
\end{table}

\section{Discussion}
\label{sec:discussion}

We delineate the boundaries of the current design. \textbf{(1) Limited atomic-operation vocabulary.} $\mathcal{S}$ covers tabletop-scale skills. Extremely contact-rich behaviors (e.g., cable insertion) may need finer decomposition than a single token, and the human-defined vocabulary injects bias: vertical bottle grasps and cabinet-handle grasps share one \texttt{grasp} token. Hierarchical or duration-parameterized tokens are a natural direction. \textbf{(2) Sensitivity to agent-side perception.} End-to-end performance depends on agent-produced boxes, so deployments pay for perception mistakes. Training-time prompt augmentation (Sec.~\ref{sec:method:objective}) only partially mitigates this, motivating uncertainty-aware prompting and closed-loop refinement. \textbf{(3) Reliance on system-level auxiliary motions.} All interaction skills are learned. Conventional planning handles only auxiliary motions (e.g., homing) and can break down under complex contact or deformable objects. \textbf{(4) Data-bound skill precision.} The precision of the atomic skills is currently bounded by the limited training data. Continual learning through agent-in-the-loop real-robot reinforcement learning is a promising direction, improving success rates and reducing retries while the system operates.

\section{Conclusions}
\label{sec:conclusion}

We have presented AR-WAM, a visual-conditioned, agent-ready world action model built on the premise that the interface between agents and manipulation policies should be spatial rather than linguistic. AR-WAM replaces language instructions with a visual grounding prompt specifying where to act and a learnable operation token specifying which skill to perform, realized by a compact 0.5B policy in which a reasoning expert externalizes task understanding into explicit, supervisable signals and dual-level foresight keeps action generation future-aware. Planning, memory, and error recovery are delegated to external agents through a model-agnostic interface of three primitives: \texttt{detect}, \texttt{execute}, and \texttt{query}. Experiments on RoboTwin 2.0, RMBench, and a real Astribot S1 platform confirm the soundness of these choices: visual prompts outperform language conditioning under identical training, off-the-shelf agents drive the policy through memory-dependent long-horizon tasks, and the compact policy transfers to real hardware with only a few demonstrations while attaining the lowest latency among compared methods. We believe such a visual, agent-native interface is the right contract between high-level agents and low-level manipulation policies.

\bibliographystyle{IEEEtran}
\bibliography{refs}

\end{document}